\documentclass{article}

\usepackage[preprint]{neurips_2026}

\usepackage[utf8]{inputenc}
\usepackage[T1]{fontenc}
\usepackage[hidelinks]{hyperref}
\usepackage{url}
\usepackage{booktabs}
\usepackage{amsfonts}
\usepackage{nicefrac}
\usepackage{microtype}
\usepackage{xcolor}
\usepackage{graphicx}
\usepackage{subcaption}
\usepackage{amsmath}
\usepackage{fontawesome5}   

\usepackage[capitalize,noabbrev,nameinlink]{cleveref}

\crefname{figure}{Fig.}{Figs.}

\crefname{snippet}{Snippet}{Snippets}
\Crefname{snippet}{Snippet}{Snippets}
\Crefname{figure}{Fig.}{Figs.}
\crefname{subfigure}{Fig.}{Figs.}
\Crefname{subfigure}{Fig.}{Figs.}

\crefname{section}{\S}{\S\S}
\Crefname{section}{\S}{\S\S}
\crefname{subsection}{\S}{\S\S}
\Crefname{subsection}{\S}{\S\S}
\crefname{subsubsection}{\S}{\S\S}
\Crefname{subsubsection}{\S}{\S\S}
\crefformat{section}{\S#2#1#3}
\crefformat{subsection}{\S#2#1#3}
\crefformat{subsubsection}{\S#2#1#3}
\crefrangeformat{section}{\S\S#3#1#4 to~#5#2#6}
\crefmultiformat{section}{\S\S#2#1#3}{ and~#2#1#3}{, #2#1#3}{ and~#2#1#3}

\usepackage{euan-macros}

\usepackage[draft]{commenting}
\declareauthor{jb}{JB}{teal}\authorcommand{jb}{comment}
\declareauthor{cds}{CDS}{purple}\authorcommand{cds}{comment}

\title{Building Better Activation Oracles}

\author{%
  Jan Bauer\thanks{Equal contribution; author order determined randomly.} \\
  MATS \\
  Gatsby Unit, UCL \\
  \And
  Celeste De Schamphelaere\footnotemark[1] \\
  MATS \\
  Ghent University \\
  \And
  Adam Karvonen \\
  Independent \\
  \And
  Niclas Luick \\
  MATS, University of Hamburg \\
  \And
  Neel Nanda \\
}

\begin{document}

\maketitle

\begin{abstract}
\emph{Activation Oracles} (AOs) are promising methods for interpreting residual stream activations.
However, current AOs face important issues, such as hallucinations and vagueness. Additionally, text-inversion confounds make them hard to evaluate.
To address this, we improve the Activation Oracle (AO) training regime in four ways: training on on-policy rollouts, improving the conversational dataset, feeding more layers, and an improvement to the injection formula. The capability improvements are marginal, but quality of life improvements are substantial. In addition, we open source the first comprehensive evaluation suite for AO quality, which we call \textbf{AObench}.
Overall, we hope that our work sets a foundation that helps improve AOs and other models in the paradigm of scalable, end-to-end interpretability.
\end{abstract}

\begin{figure}[h]
  \centering
  \includegraphics[width=0.85\linewidth]{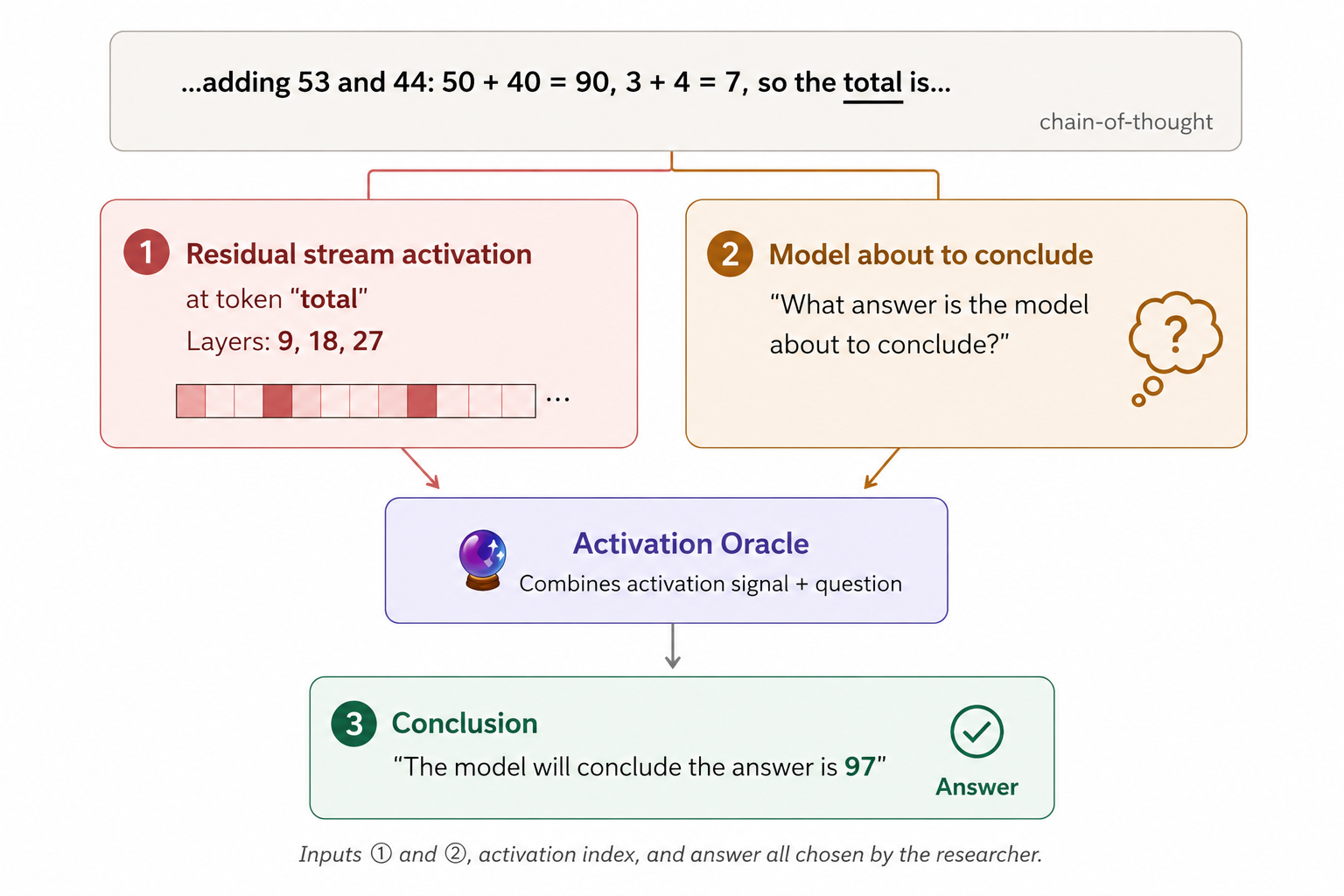}
  \caption{\textbf{Activation Oracle overview}. The Oracle receives residual-stream activations and a natural-language question, then produces an answer about the model state represented by those activations.}
  \label{fig:ao-overview}
\end{figure}

\section{Introduction}

Activation Oracles (henceforth AOs) \citep{karvonen2025activation} are finetuned LLMs that can receive the original LLM's activations as input and answer natural language questions about them.

However, current AOs face several issues that make them hard to use, such as hallucinations, vagueness, and a lack of verifiable faithfulness \citep{jakkli2026current}. Additionally, text-inversion confounds (where a model can match a hypothetical "true" oracle's performance by simply reconstructing the surrounding text from an activation and answer purely from this reconstructed text) make them hard to evaluate.

The standard AO training recipe comprises three components: the LatentQA conversational dataset \citep{pan2024latentqa}, a suite of binary classification tasks, and a self-supervised past/future-lens objective trained on FineWeb, in which the AO predicts tokens before or after a sequence of activations. We identify issues with LatentQA as a dataset, and the FineWeb past/future-lens task as a training objective. Activations are fed via a norm-matched injection formula, after the second transformer layer \citep{karvonen2025activation}.

We propose to partially alleviate these problems by constructing a better conversational dataset, feeding activations from multiple layers and multiple token positions, training on on-policy data and increasing the magnitude of the activation injection.

We find that these changes (particularly the new conversational dataset) produce consistent improvements: in both quantitative evaluation and qualitative testing, the resulting AO scores higher overall on evaluations, follows instructions better, hallucinates less, and is substantially less \emph{vague} than the original AO checkpoint. To enable further work in this direction, we release \textbf{AObench}, the first comprehensive evaluation suite for AO quality, designed to measure what an ideal AO should be good at while attempting to remain robust to text-inversion confounds, all while targeting their major issues.\footnote{Code: \url{https://github.com/japhba/activation_oracles}. Models and datasets: \url{https://huggingface.co/collections/ceselder/building-better-activation-oracles}}

Conversely, we find narrow post-training on the tasks from \citet{ivanova2026test} consistently fails to exceed simple linear-probe performance.

We see AOs as part of the emerging paradigm of scalable, end-to-end interpretability \citep{steinhardt2025scalable, pan2024latentqa, karvonen2025activation, choi2025usermodeling, huang2025predictive, li2026traininglanguagemodelsexplain, frasertaliente2026nla}: training models on self-supervised objectives to map model internals to natural-language explanations. We believe the largest wins for AOs will likely come from improvements to the unsupervised training task itself; our contributions take a step in this direction, but we expect substantial further headroom if a scalable task which increases capability can be found.

\section{Issues with current Activation Oracles}

\citet{jakkli2026current} demonstrate scenarios in which AOs are hard to work with. We focus on addressing two of their issues: \textbf{hallucinations}, where the AO outputs false information, and \textbf{vagueness}, where the AO output is generic (and therefore unfalsifiable) and does not answer the user's question. 

They additionally highlight the problem of \textbf{text inversion}: the model can succeed simply by inferring the surrounding text and answering from that reconstruction, just as any black-box oracle could; this is a major frustration in evaluating AOs.

\section{Improving Activation Oracle training}

\subsection{A better conversational dataset}
To enable the Activation Oracle to answer natural language questions, a dataset consisting of questions and answers about activations is needed. To this end, the original paper used \emph{LatentQA}. \citep{pan2024latentqa}

However, we found that this dataset was of low quality, likely incentivizing vagueness. We isolate three issues:

\begin{itemize}
    \item The model is given a complicated prompt, and then a specific question is asked about this prompt. We think the answers to the questions LatentQA poses are often not easily retrievable from activations, which makes it a difficult task for the AO, not incentivizing much beyond text inversion, and may even directly incentivize hallucinations/guessing if the relevant info is not present.
    \item The questions are not about on-policy data, but about specifics of a user prompt: this does not target the model’s internal reasoning.  
    \item It was generated by o1, a now outdated model.
\end{itemize}

We constructed a new conversational dataset that attempts to address all of these concerns. Because we don’t want the questions to be trivially answerable from adjacent tokens (text inversion), we construct QA pairs as follows: a separate LLM (Sonnet 4.6) is given the target model’s chain-of-thought (CoT), and is instructed to split the chain-of-thought into a prefix and suffix, and to write a question about the suffix. It is instructed to do this in a way such that the question is hard to answer purely from the text of the prefix (i.e. to avoid text inversion), but plausibly answerable from the prefix’s activations (solvability). \footnote{You can explore our dataset here: \url{https://huggingface.co/datasets/ceselder/cot-oracle-convqa-chunked-sonnet}}

\begin{figure}[!h]
  \centering
  \includegraphics[width=0.75\linewidth]{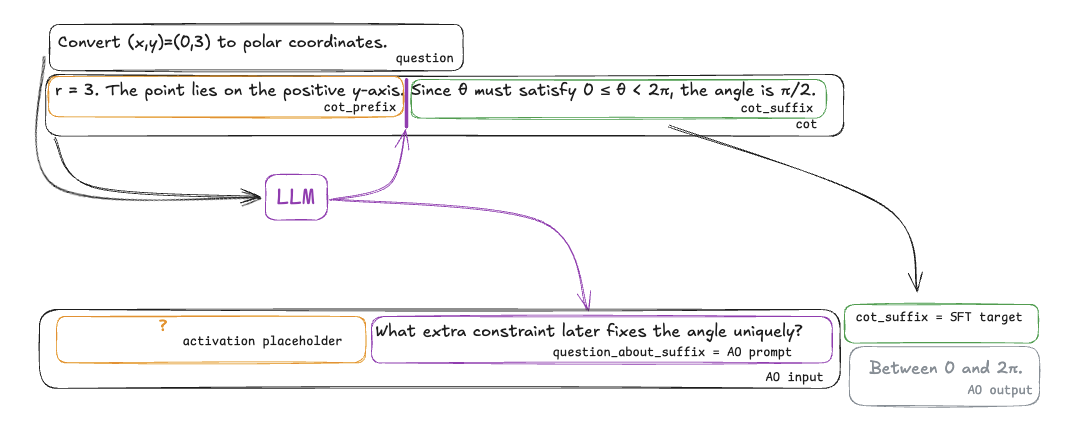}
  \caption{\textbf{How our conversational dataset is constructed.} A language model is asked to split a chain-of-thought and to produce a question about the split suffix that is plausibly answerable from the split prefix.}
  \label{fig:ablation-convqa}
\end{figure}

\begin{figure}[!h]
  \centering
  \includegraphics[width=0.5\linewidth]{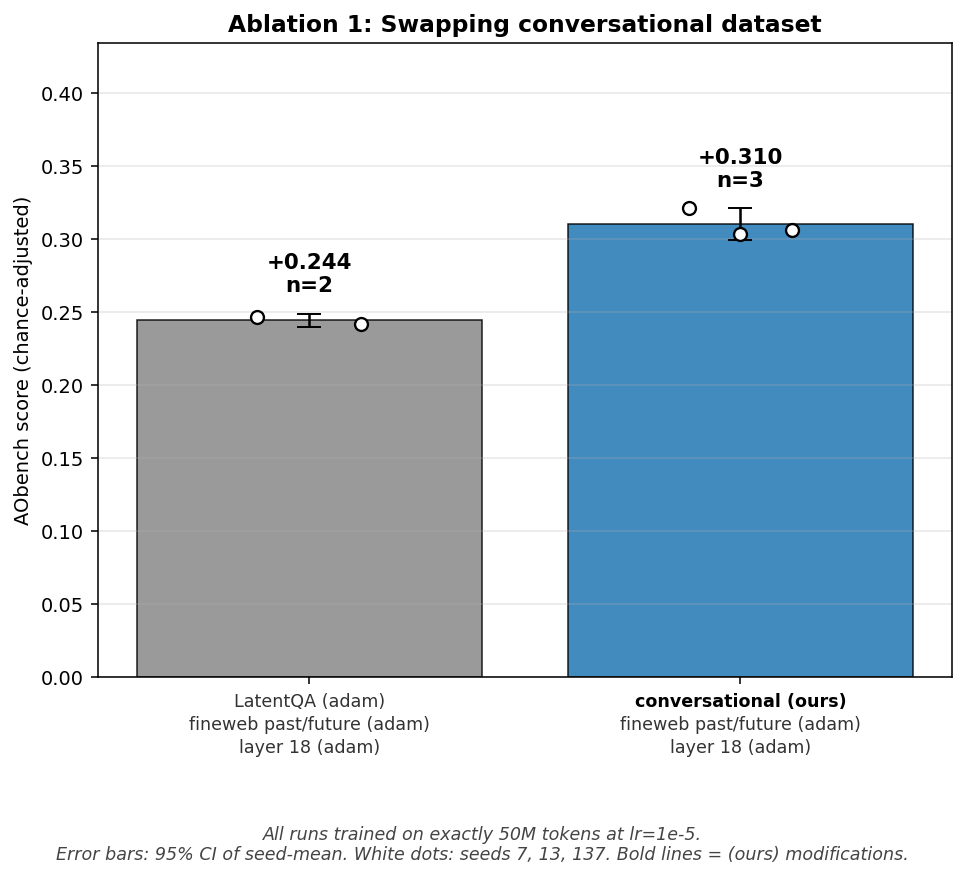}
  \caption{\textbf{Conversational dataset swap, isolated.} Replacing only LatentQA \citep{pan2024latentqa} with our conversational dataset (leaving past/future-lens corpus and layer choice fixed) improves chance-adjusted AObench score from $+0.244$ to $+0.310$ ($n=3$ seeds). This is the single largest step in our recipe.}
  \label{fig:ablation-convqa}
\end{figure}

We ablate the effect of this task in \cref{fig:ablation-convqa} by replacing only LatentQA in Adam's recipe (leaving everything else the same) and notice a significant uplift, across the board on our AObench evaluations. We find that the responses are more specific and the resulting model is less vague, and responds better to instructions.

\subsection{Layer choice/feeding multiple layers to the AO}

Adam originally fed activations randomly selected from either layer 25\%, 50\% or 75\% of total model depth. Since most features live around the 55-80\% layer ranges, we suspected a layer sweep could be important. Indeed, we find that AO performance peaks at layer 22 (62\%). Feeding 5 contiguous layers from layer 21-25 causes further uplift. Interestingly, the largest uplift is on model diffing tasks. We’d like to point out that training a multi-layer Activation Oracle can cause an increase in training time due to longer context, and that most gains can be had by simply choosing a layer at ~65\% depth (though this may differ per model, and per application).

\begin{figure}[!h]
  \centering
  \includegraphics[width=0.5\linewidth]{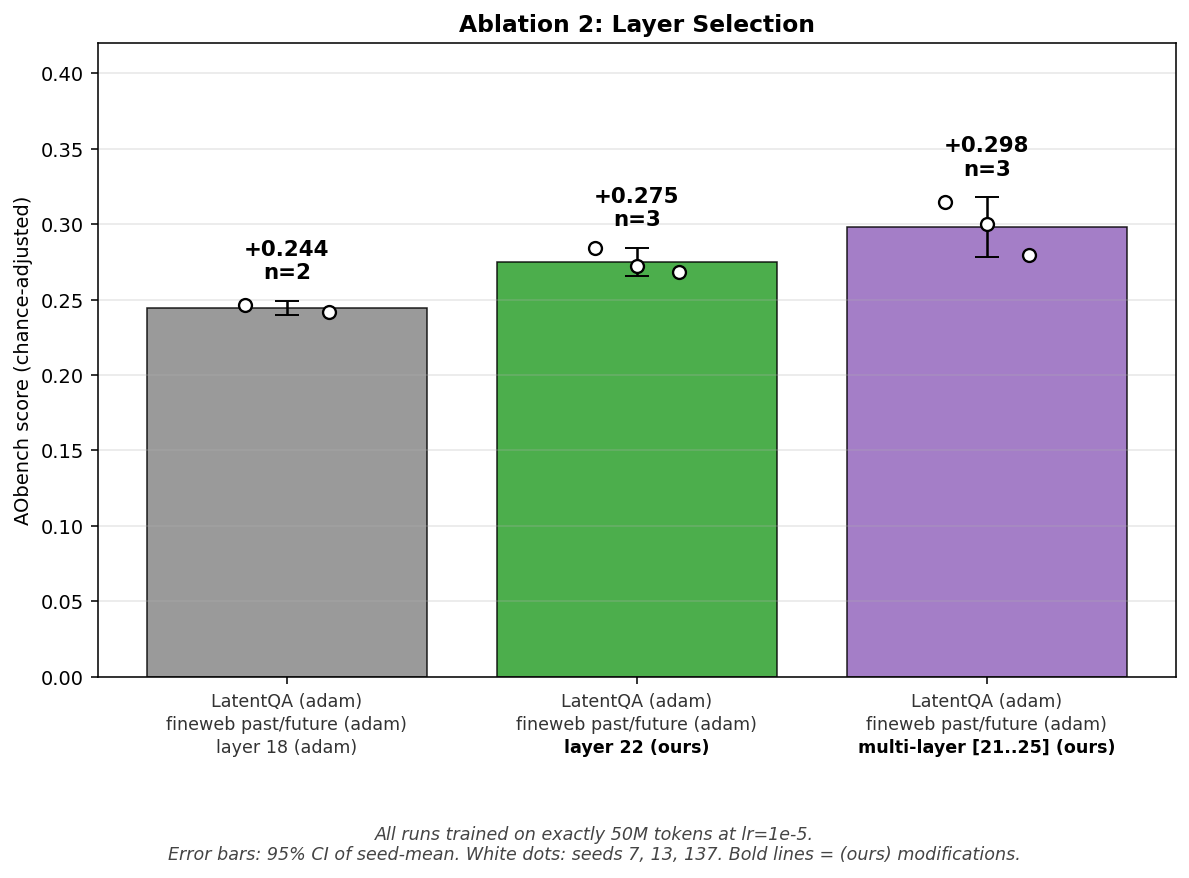}
  \caption{\textbf{Layer sweep.} Layer 22 causes improved performance over Layer 18 (+0.025 on AObench), and 5 contiguous layers even further still (+0.05 on AObench)}
  \label{fig:ablation-layersweep}
\end{figure}

\subsection{Training on on-policy data}

To train Activation Oracles, we need scalable unsupervised training tasks. A common way to achieve this is to predict past and/or future tokens from the activations, known as past or future-lens. This requires some data to source activations from, from which then to predict tokens.

\begin{figure}[!h]
  \centering
  \includegraphics[width=0.5\linewidth]{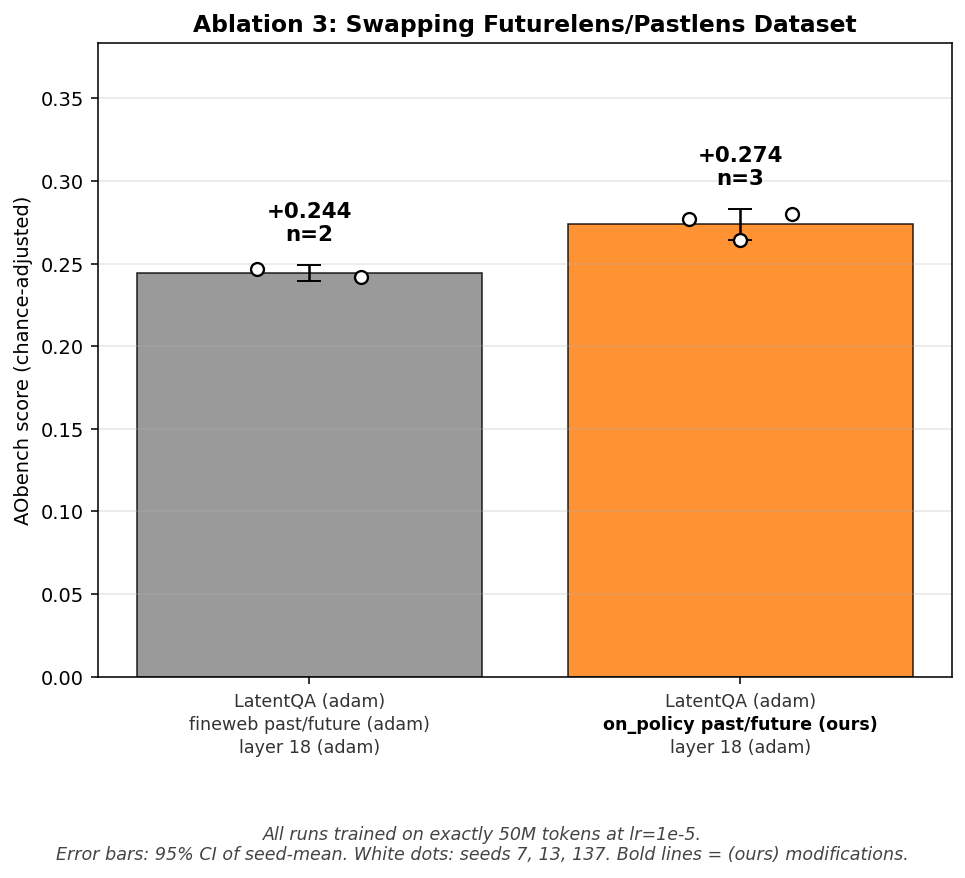}
  \caption{\textbf{On-policy data.} Replacing only the past/future-lens corpus from FineWeb to on-policy chain-of-thought rollouts improves chance-adjusted AObench score from $+0.244$ to $+0.274$ ($n=3$ seeds), a smaller effect than the conversational swap.}
  \label{fig:ablation-onpolicy}
\end{figure}

Adam’s original paper only used pre-training data \citep{penedo2024fineweb}. However, this has a problem: to predict future tokens in pre-training data, you don't necessarily need to know much about what the model is thinking, just what the prior text is. 

We think that the on-policy data we use (i.e., generations from the model we are trying to interpret) are better training data because we hypothesize it to be a more solvable task, by virtue of targeting what the model is actually representing in its activations. Further, we will in practice use the AO on a model in an on-policy setting, e.g. for studying agent traces. While the above explanation is plausible, we only notice minor uplift in evaluations.

We swap FineWeb for on-policy corpus in \cref{fig:ablation-onpolicy}; this change yields a small but measurable improvement on AObench (\textasciitilde$+0.03$). While the above explanation is plausible, the magnitude of the effect is modest.

\subsection{Injection strength}
Natural Language Autoencoders (NLAs) \citep{frasertaliente2026nla} inject their activations by replacing the token embedding entirely, and using a fixed scalar. We use additive, norm-matched injection after the second transformer layer following \cite{karvonen2025activation}. 

We do not have a formal ablation for this, but on Qwen3-8B, every run that did NLA-style injection performed significantly worse than Adam’s formula. 

\cite{frasertaliente2026nla} sweep their injection strength and claim that this is a quite sensitive hyperparameter. We did the same starting from Adam’s formula, and found that increasing the injection strength marginally increases performance. This difference may look small, and indeed it is, but in hallucinations it is considerable (79\% to 85\%), which is particularly important, so we do recommend carefully choosing your hyperparameter value here.

\begin{figure}[h]
  \centering
  \includegraphics[width=0.5\linewidth]{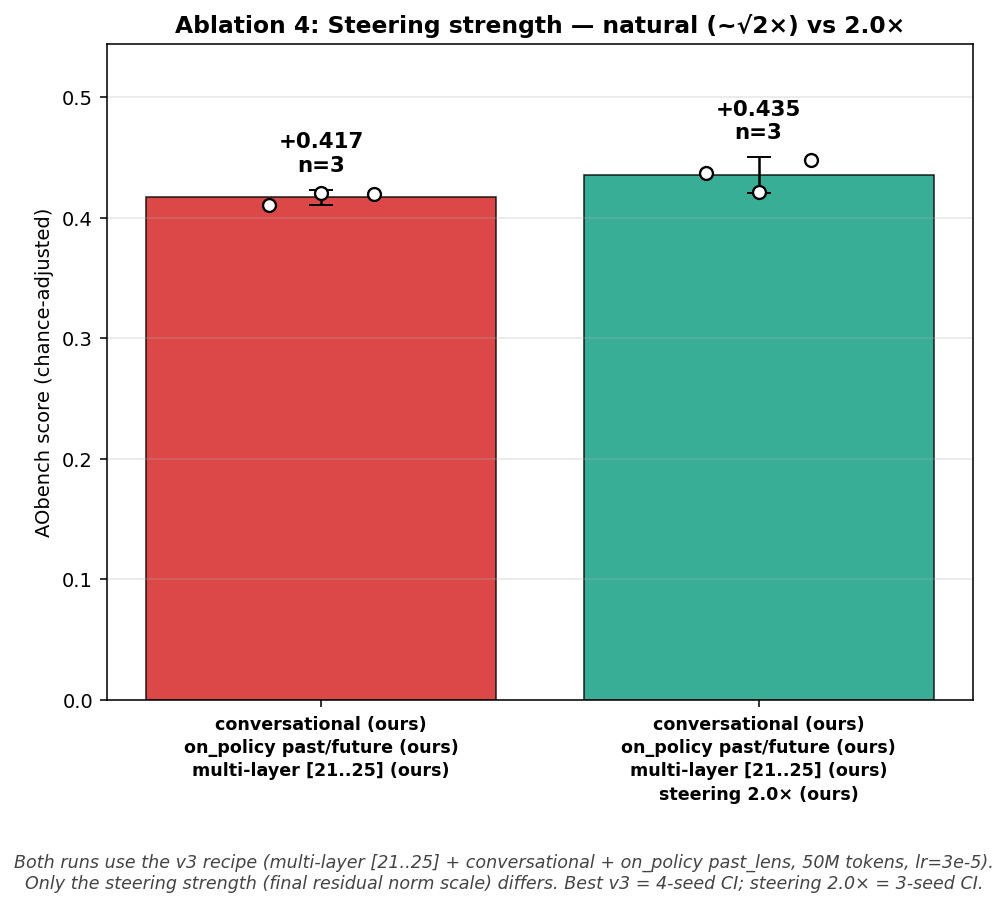}
  \caption{We ablate steering strength and find it marginally increases performance.}
  \label{fig:ablation-steering}
\end{figure}

Our hypothesis why injecting on the second layer does better than replacing the embedding, is that the first residual stream layer has a very small cosine similarity to previous layers, a property unique to the first layer. After the first layer, cosine similarities remain fairly stable layer to layer. Because of this, it’s plausible that injecting after the second layer, when the residual stream lies in the ``correct basis'', would work better. (\cite{karvonen2025activation} reaches similar conclusions). The reason a stronger injection strength might do better is that language models have a strong prior to weight tokens sort of equally, and that it’s rare that one token is load bearing for the entire explanation. Language model priors can be hard to overcome, so manually enforcing a stronger norm for the activation can help overcome this.

\section{Results}\label{sec:results}

We constructed AObench \textbf{with the aim of measuring what an ideal Activation Oracle should be good at}. The benchmark is a work in progress, but we recommend it as a starting point for evaluating new Activation Oracles. It targets the main frustrations identified by \citet{jakkli2026current} and reuses several model organisms from \citet{karvonen2025activation}; full per-task results and prompts are reported in \cref{app:aobench}. Concretely, \emph{vagueness} evaluates whether the oracle's description of the model's reasoning is concrete and problem-specific, and \emph{hallucination} evaluates whether the oracle invents specific but unsupported details about the model's reasoning.

We perform controlled ablations, starting from \citet{karvonen2025activation} where we apply each of our changes. All runs are trained for exactly 50M tokens at matched learning rates, with multiple seeds where compute allowed.\footnote{Our checkpoints are slightly undertrained (50M vs 65M tokens) relative to \citet{karvonen2025activation}; we expect further training to improve absolute performance, but the relative ordering of recipes is already stable at our token budget (see \cref{app:hillclimb}).} The full recipe improves chance-adjusted AObench score from $+0.244$ (Adam baseline) to $+0.435$, with the conversational dataset swap alone accounting for the largest single jump ($+0.244 \to +0.310$).

\begin{figure}[htbp]
  \centering
  \includegraphics[width=0.95\linewidth]{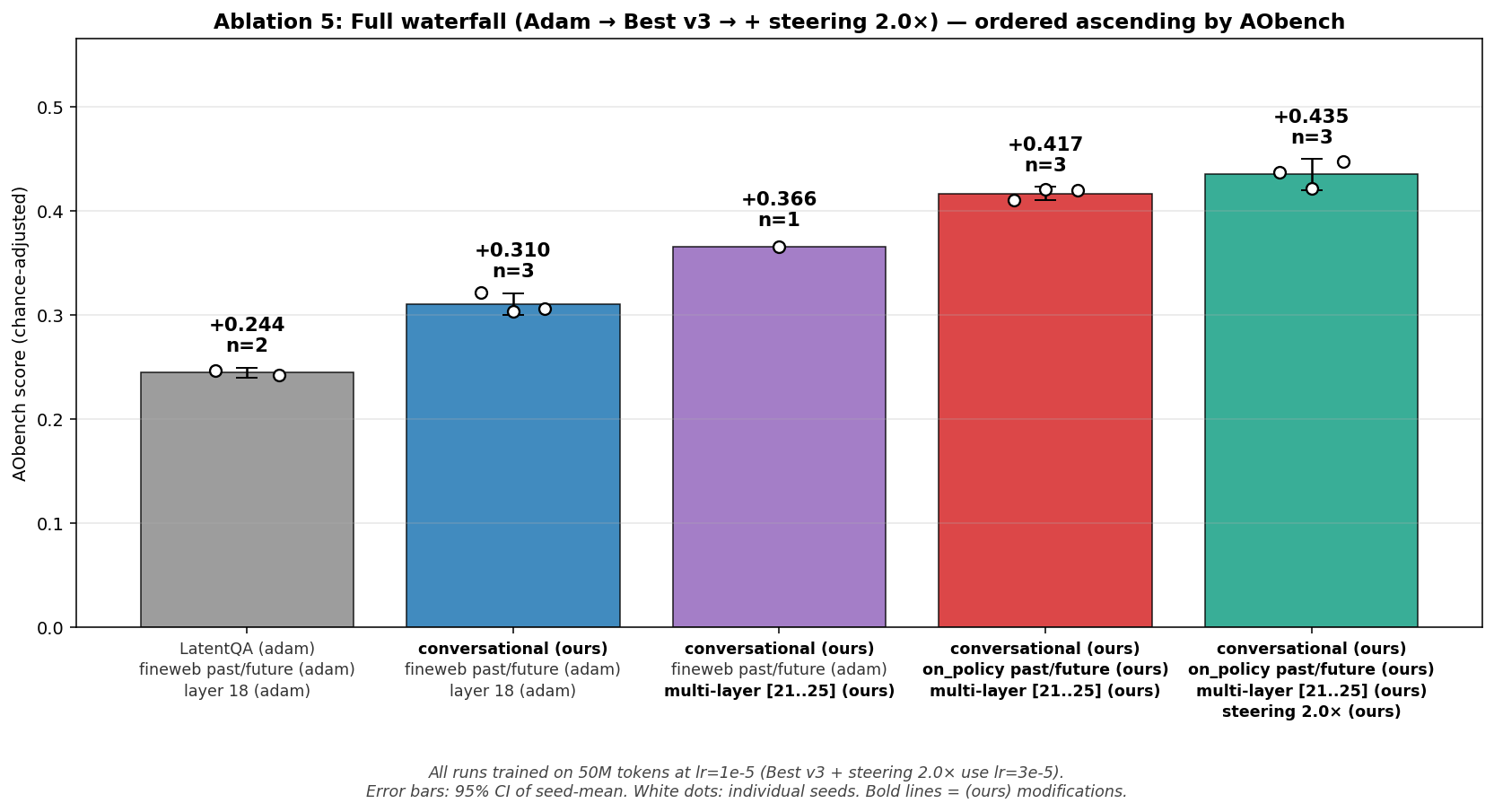}
  \caption{\textbf{AObench ablation ladder.} Each bar adds one of our interventions on top of the previous recipe. The conversational dataset swap (blue) drives the largest single-step improvement; multi-layer extraction and on-policy past/future-lens data each contribute additional uplift, and a 2$\times$ injection-strength tweak yields a final small gain. All runs trained on 50M tokens; error bars show 95\% CI of seed mean.}
  \label{fig:ablation-waterfall}
\end{figure}

\begin{figure}[htbp]
  \centering
  \begin{subfigure}[t]{0.49\linewidth}
    \centering
    \includegraphics[width=\linewidth]{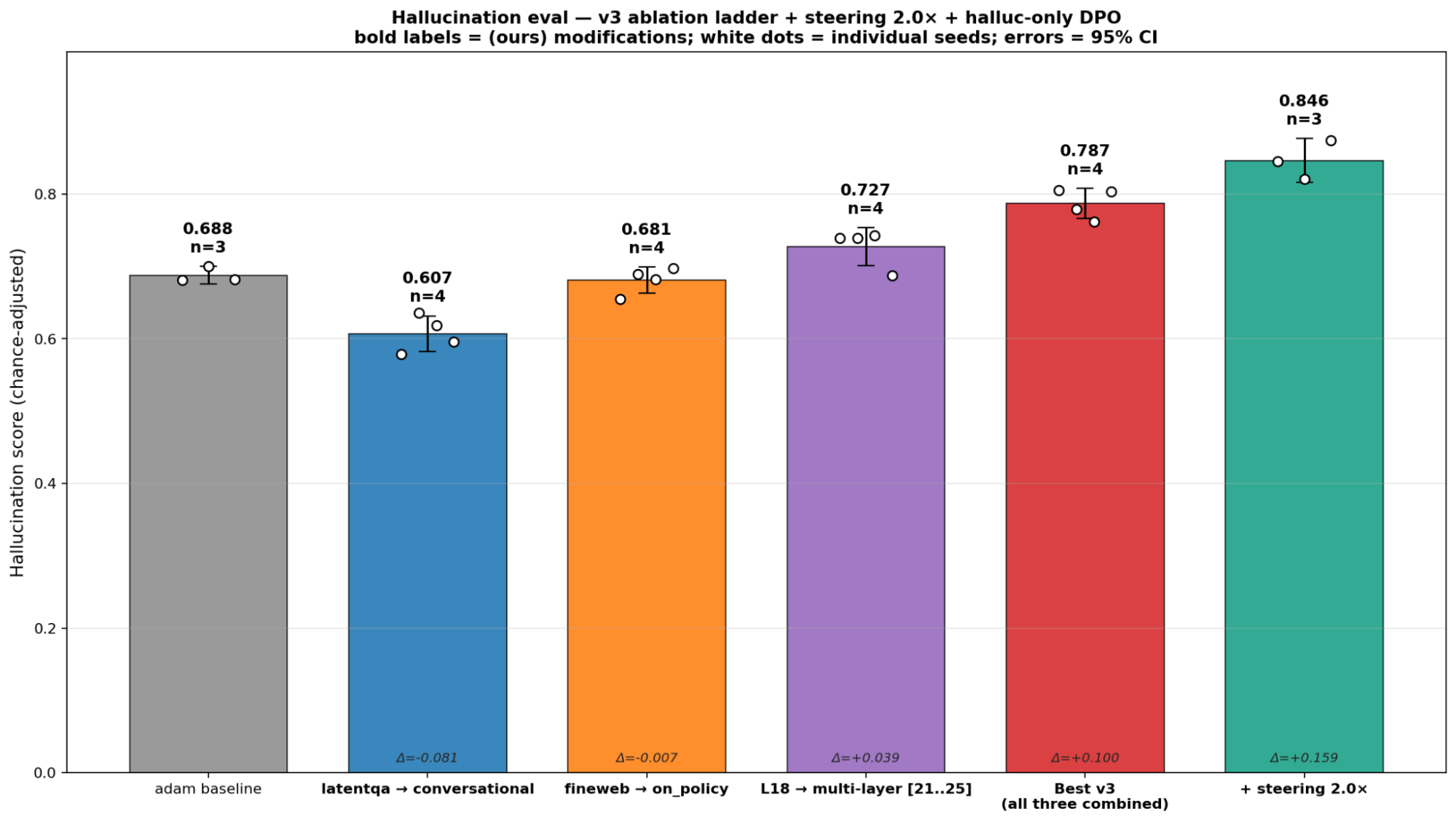}
    \label{fig:hallucination-ladder}
  \end{subfigure}\hfill
  \begin{subfigure}[t]{0.49\linewidth}
    \centering
    \includegraphics[width=\linewidth]{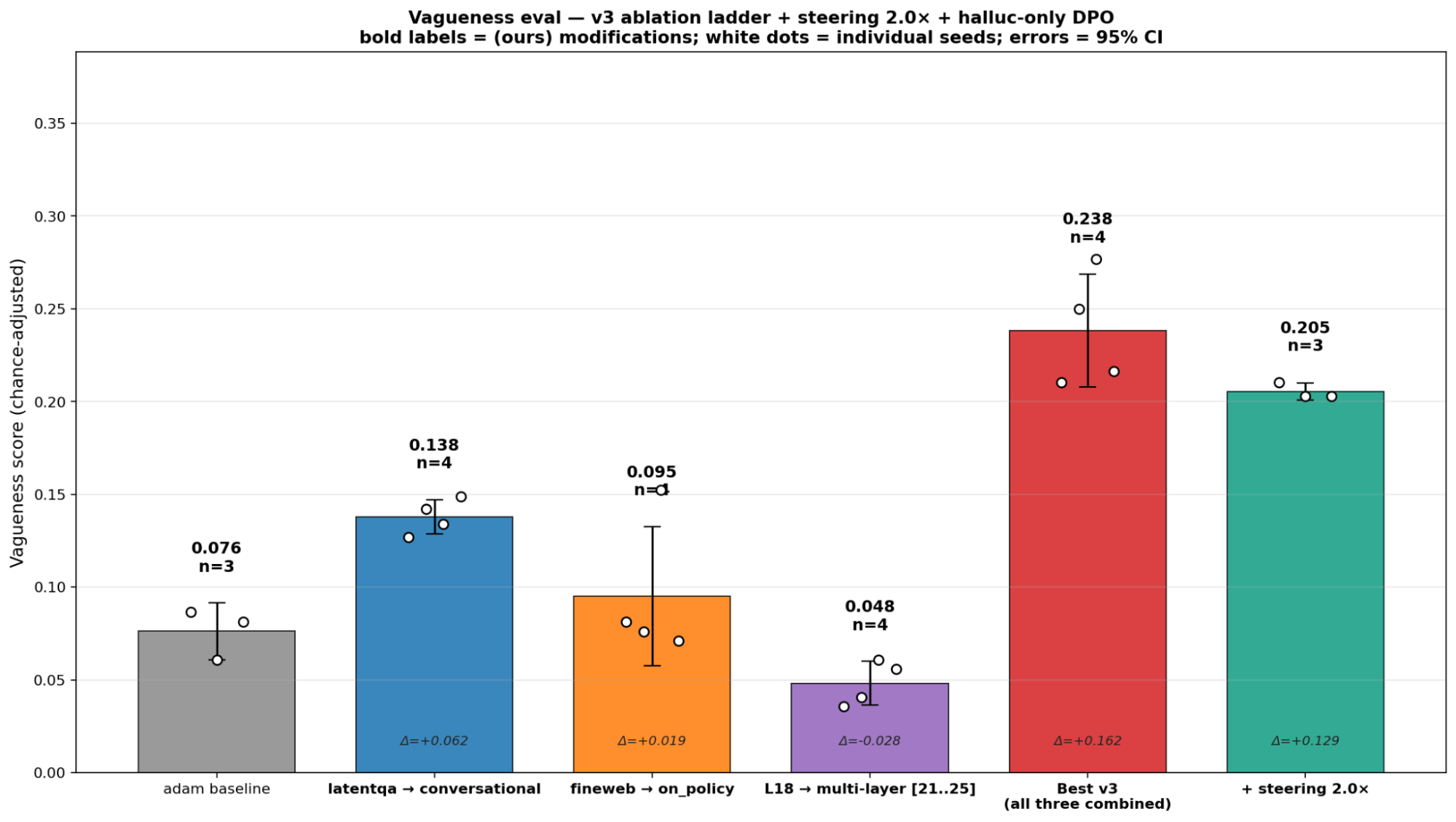}
    \label{fig:vagueness-ladder}
  \end{subfigure}
  \caption{\textbf{Hallucination and vagueness across the ablation ladder.} Each bar adds one of our interventions on top of the previous recipe; error bars are 95\% CI of seed mean.}
  \label{fig:hall-vague-ladder}
\end{figure}

\paragraph{Hallucination and vagueness.}
\Cref{fig:hall-vague-ladder} decomposes the ablation along the two axes most highlighted by \citet{jakkli2026current}. It's important to note that hallucination score initially increases; we attribute this to our conversational data training the model to make more specific claims, which makes it more likely for a hallucination to be counted as such. After accounting for this, the increase is monotonic (68.8\% to 84.6\%). Vagueness improves substantially in the full recipe relative to the original AO ($0.076 \to 0.205$ chance-adjusted), with the conversational dataset and multi-layer interventions contributing the majority of the gain.

\paragraph{Caveats.}
The on-policy past/future-lens ablation is not perfectly clean, because our conversational dataset is also constructed from on-policy chain-of-thought rollouts; some of what the on-policy lens contributes may already be expressible through the conversational data. We also caution that AOs remain difficult to evaluate: LLM judges are noisy on open-ended outputs, and several of our metrics are sensitive to prompt phrasing (\cref{app:eval-practice}). With those caveats, the ordering in \cref{fig:ablation-waterfall} has been stable across seeds and minor recipe variations.

\section{Outlook}

After substantial work on AOs, we believe they are a useful interpretability technique, but aren't the best tool in all circumstances. They are best used for complex open-ended questions about activations, for instance, making sense of why the model backtracked. We expect them to be particularly valuable for interpreting latent-reasoning models or any setting where substantial computation happens within a single forward pass and is therefore inaccessible to chain-of-thought monitoring. However, even with our improvements, clear limitations remain. First, they still hallucinate frequently, though this generally improves with the amount of activations supplied and uncertainty can be estimated by resampling (\cref{app:eval-practice}). Second, in many settings (but not all settings) it is possible to just read the chain-of-thought directly and arrive at the same insight as the AO.

Still, we think there may be significant room for improvement by scaling up the conversational data we used, both in amount and kind. A second route is to include more narrow tasks in a ``post-training'' stage, though this did not yield improvements at our AO's current capability margin. Another exciting path forward is to come up with more evaluations that target something an ideal AO could plausibly solve, while being robust to text inversion concerns. If such tasks are scalable, they can be used for training as well.

More broadly, we think the idea of a scalable meta-model for activations is a promising interpretability agenda, that may scale well with model capabilities. We do not think the failures of current AOs are reason to rule out this approach.

We are particularly excited about Natural Language Autoencoders (NLAs) \citep{frasertaliente2026nla}, and expect them to ultimately be a better way to pretrain AOs than pastlens/futurelens, since they offer a principled way to bootstrap natural-language ground truth from activations. Our observations regarding LatentQA however, remain applicable. Indeed, the original NLA paper performs its conversational finetune using LatentQA. We expect it would benefit substantially from either (i) a conversational dataset constructed along the solvability and targetedness principles introduced here, or (ii) constructing QA pairs from topics that appear in a consensus over many NLA rollouts on the same activation; essentially bootstrapping conversational solvability from bootstrapped ground truth.

We refer the reader interested in further advice for training AOs to \cref{app:hillclimb}.

\section*{Contributions}
Jan Bauer and Celeste De Schamphelaere contributed equally and carried out all experiments and writing. Niclas Luick came up with the idea of multilayer activation oracles, did the initial development and experimentation, and did initial exploration on quantifying uncertainty via consensus sampling. Adam Karvonen provided mentorship and guidance. Neel Nanda provided senior mentorship. 

\section*{Acknowledgments}
This work was conducted as part of the ML Alignment \& Theory Scholars
(MATS) program (cohort 10.0). We thank the MATS program for funding and compute.

\newpage
\appendix


\bibliographystyle{plainnat}
\bibliography{refs}

@article{karvonen2025activation,
  author     = {Karvonen, Adam and Chua, James and Dumas, Cl{\'e}ment and Fraser-Taliente, Kit and Kantamneni, Subhash and Minder, Julian and Ong, Euan and Sen Sharma, Arnab and Wen, Daniel and Evans, Owain and Marks, Samuel},
  title      = {Activation Oracles: Training and Evaluating {LLMs} as General-Purpose Activation Explainers},
  journal    = {arXiv preprint arXiv:2512.15674},
  year       = {2025},
  eprint     = {2512.15674},
  eprinttype = {arxiv},
  url        = {https://arxiv.org/abs/2512.15674},
  googlescholar = {https://scholar.google.com/scholar?cluster=10298672171677107208},
}

@article{cywinski2025elicit,
  author     = {Cywi{\'n}ski, Bartosz and Ryd, Emil and Rajamanoharan, Senthooran and Nanda, Neel},
  title      = {Towards Eliciting Latent Knowledge from {LLM}s with Mechanistic Interpretability},
  journal    = {arXiv preprint arXiv:2505.14352},
  year       = {2025},
  eprint     = {2505.14352},
  eprinttype = {arxiv},
  url        = {https://arxiv.org/abs/2505.14352}
}

@misc{steinhardt2025scalable,
  author       = {Steinhardt, Jacob},
  title        = {Scalable End-to-End Interpretability},
  howpublished = {LessWrong},
  year         = {2025},
  month        = dec,
  url          = {https://www.lesswrong.com/posts/qkhwh4AdG7kXgELCD/scalable-end-to-end-interpretability},
  googlescholar = {https://scholar.google.com/scholar?q=%22Scalable+End-to-End+Interpretability%22+Steinhardt},
}

@article{huang2025predictive,
  author     = {Huang, Vincent and Choi, Dami and Johnson, Daniel D. and Schwettmann, Sarah and Steinhardt, Jacob},
  title      = {Predictive Concept Decoders: Training Scalable End-to-End Interpretability Assistants},
  journal    = {arXiv preprint arXiv:2512.15712},
  year       = {2025},
  eprint     = {2512.15712},
  eprinttype = {arxiv},
  url        = {https://arxiv.org/abs/2512.15712},
  googlescholar = {https://scholar.google.com/scholar?cluster=7648866656024750732},
}

@misc{choi2025usermodeling,
  author       = {Choi, Dami and Huang, Vincent and Schwettmann, Sarah and Steinhardt, Jacob},
  title        = {Scalably Extracting Latent Representations of Users},
  howpublished = {Transluce, \url{https://transluce.org/user-modeling}},
  year         = {2025},
  month        = nov,
  googlescholar = {https://scholar.google.com/scholar?cluster=10074423253601403691},
}

@article{penedo2024fineweb,
  author     = {Penedo, Guilherme and Kydl{\'\i}{\v{c}}ek, Hynek and Ben allal, Loubna and Lozhkov, Anton and Mitchell, Margaret and Raffel, Colin and Von Werra, Leandro and Wolf, Thomas},
  title      = {The {FineWeb} datasets: Decanting the web for the finest text data at scale},
  journal    = {arXiv preprint arXiv:2406.17557},
  year       = {2024},
  eprint     = {2406.17557},
  eprinttype = {arxiv},
  url        = {https://arxiv.org/abs/2406.17557},
  googlescholar = {https://scholar.google.com/scholar?cluster=12077055439257234910},
}

@misc{li2026traininglanguagemodelsexplain,
      title={Training Language Models to Explain Their Own Computations},
      author={Belinda Z. Li and Zifan Carl Guo and Vincent Huang and Jacob Steinhardt and Jacob Andreas},
      year={2026},
      eprint={2511.08579},
      archivePrefix={arXiv},
      primaryClass={cs.CL},
      url={https://arxiv.org/abs/2511.08579},
      googlescholar = {https://scholar.google.com/scholar?cluster=10978110331633993734},
}

@article{pan2024latentqa,
  author     = {Pan, Alexander and Chen, Lijie and Steinhardt, Jacob},
  title      = {{LatentQA}: Teaching {LLMs} to Decode Activations Into Natural Language},
  journal    = {arXiv preprint arXiv:2412.08686},
  year       = {2024},
  eprint     = {2412.08686},
  eprinttype = {arxiv},
  url        = {https://arxiv.org/abs/2412.08686},
  googlescholar = {https://scholar.google.com/scholar?cluster=14023675163405251513},
}

@article{frasertaliente2026nla,
  author={Fraser-Taliente, Kit and Kantamneni, Subhash and Ong, Euan and Mossing, Dan and Lu, Christina and Bogdan, Paul C. and Ameisen, Emmanuel and Chen, James and Kishylau, Dzmitry and Pearce, Adam and Tarng, Julius and Wu, Alex and Wu, Jeff and Zhang, Yang and Ziegler, Daniel M. and Hubinger, Evan and Batson, Joshua and Lindsey, Jack and Zimmerman, Samuel and Marks, Samuel},
  title={Natural Language Autoencoders Produce Unsupervised Explanations of LLM Activations},
  journal={Transformer Circuits Thread},
  year={2026},
  url={https://transformer-circuits.pub/2026/nla/index.html}
}

@misc{jakkli2026current,
  author       = {Jakkli, Arya and Rajamanoharan, Senthooran and Nanda, Neel},
  title        = {Current activation oracles are hard to use},
  howpublished = {LessWrong},
  year         = {2026},
  month        = mar,
  url          = {https://www.lesswrong.com/posts/LXQBcztrWKhtcgQfJ/current-activation-oracles-are-hard-to-use},
  googlescholar = {https://scholar.google.com/scholar?q=%22Current+activation+oracles+are+hard+to+use%22+Jakkli},
}

@misc{ivanova2026test,
  author       = {Ivanova, Daria and Tyagi, Riya and Engels, Josh and Nanda, Neel},
  title        = {Test your best methods on our hard {C}o{T} interp tasks},
  howpublished = {LessWrong},
  year         = {2026},
  month        = mar,
  url          = {https://www.lesswrong.com/posts/tDJWZLQNN7poqCwKa/test-your-best-methods-on-our-hard-cot-interp-tasks},
  googlescholar = {https://scholar.google.com/scholar?q=%22Test+your+best+methods+on+our+hard+CoT+interp+tasks%22+Ivanova},
}

\newpage

\crefalias{section}{appendix}
\crefalias{subsection}{appendix}
\crefalias{subsubsection}{appendix}

\section{Appendix}

\subsection{Practical notes on evaluating Activation Oracles}
\label{app:eval-practice}

While building AObench, we encountered several measurement issues that materially affect reported AO performance. We document them here because they apply to any AO evaluation and are easy to get wrong.

\paragraph{Use AUC, not accuracy, for binary classification tasks.}
\citet{jakkli2026current} reported near-chance AO performance on tasks such as sycophancy detection. We find that this is largely a calibration artifact: Qwen-based AOs frequently default to answering ``No'' regardless of the question, which makes fixed-threshold accuracy near-chance but leaves a strong signal in the difference between the ``Yes'' and ``No'' token logits. On a sycophancy-from-CoT task, the original AO scores 0.50 accuracy but 0.83 ROC AUC (\cref{fig:auc-vs-acc}). AUC is also markedly less sensitive to prompt phrasing than accuracy is; we recommend it as the default metric on yes/no AObench items.

\begin{figure}[htbp]
  \centering
  \includegraphics[width=0.75\linewidth]{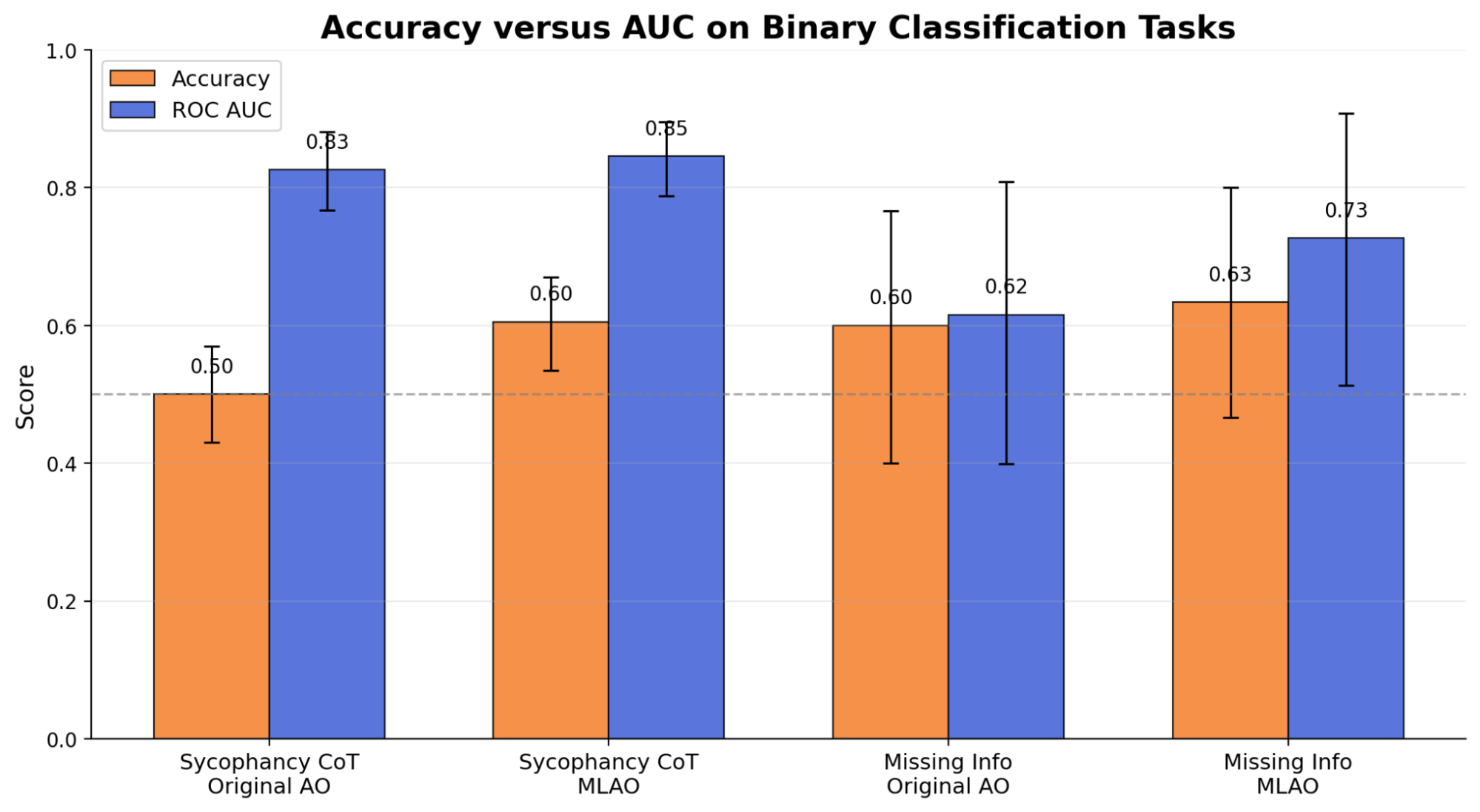}
  \caption{\textbf{Accuracy underestimates AO capability on Yes/No tasks.} Qwen-based AOs default to ``No'' on several binary-classification items, which suppresses accuracy without affecting the Yes/No logit margin. ROC AUC is much higher and more stable across phrasings.}
  \label{fig:auc-vs-acc}
\end{figure}

\paragraph{Sweep the AO's context window when comparing to black-box baselines.}
Many open-ended AO tasks (e.g., ``why is the model about to backtrack?'') concern information spread across tens of tokens of internal computation. Restricting the AO to the final activation alone consequently produces misleadingly weak performance. On a Qwen3-8B backtracking task, the original AO scores $1.26/5$ mean correctness given only the final-token activation, but $2.10/5$ given the last 50 tokens, above the black-box baseline of asking Qwen3-8B the same question with full text context (\cref{fig:backtracking-context}). We therefore report AObench results at a context window of $\geq 20$ tokens by default, and recommend the same when comparing AOs to text-only baselines.

\begin{figure}[htbp]
  \centering
  \includegraphics[width=0.7\linewidth]{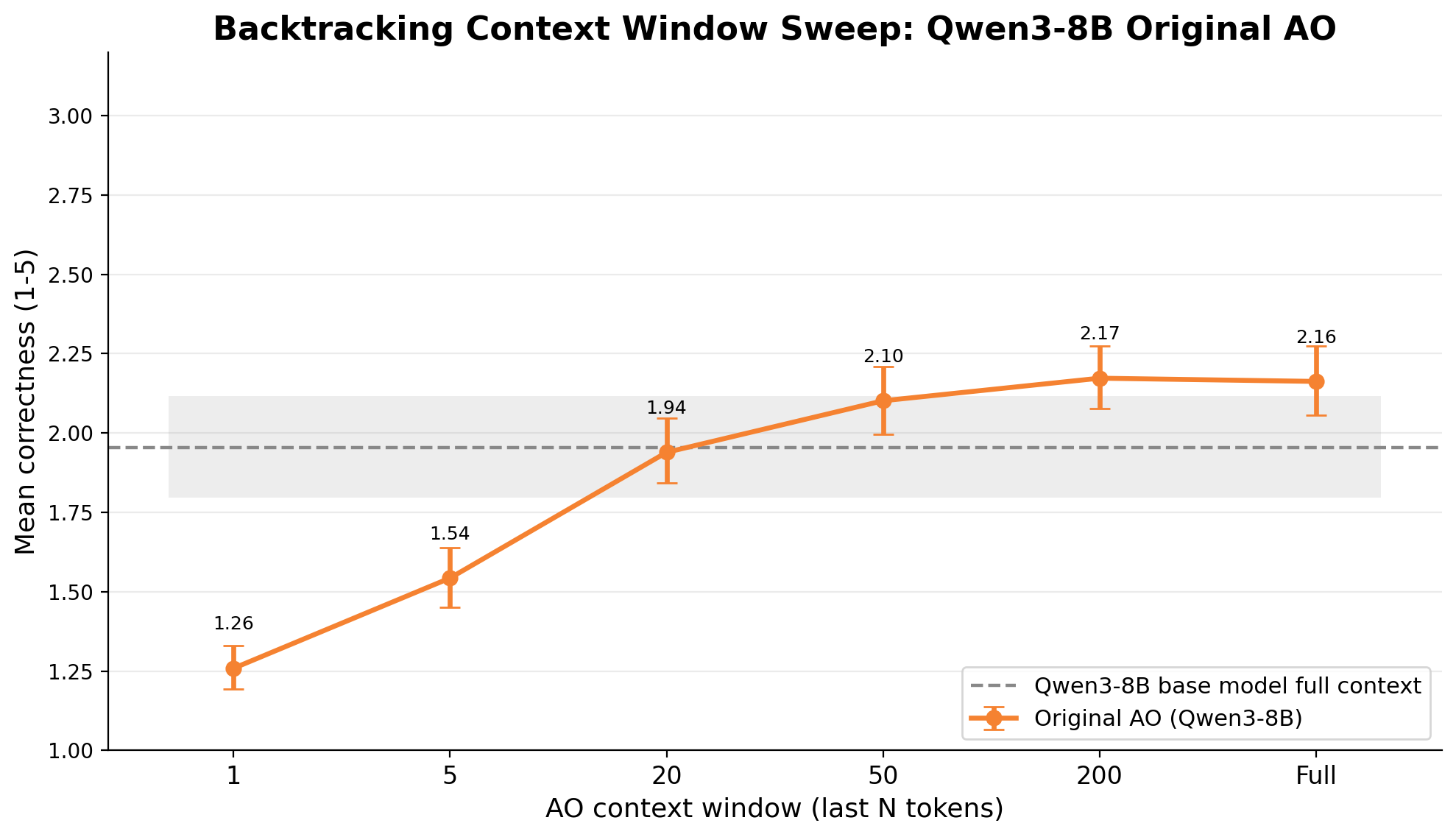}
  \caption{\textbf{Backtracking accuracy vs.\ AO context window.} Performance rises steeply with the number of activation positions supplied. The AO matches the black-box baseline at $\sim$20 tokens and exceeds it at 50.}
  \label{fig:backtracking-context}
\end{figure}

\paragraph{Consensus sampling materially mitigates open-ended hallucination.}
For free-form AO answers, a simple inference-time strategy is to sample $k$ completions at non-zero temperature and retain only those for which at least a fraction $\tau$ of samples agree. On the taboo secret-word task, unfiltered single-token accuracy is $46.6\%$; requiring consensus $\geq 0.8$ over $k=10$ samples retains $19.4\%$ of items at $94.3\%$ precision, with a smooth precision/recall trade-off (\cref{fig:consensus}). This is a cheap, training-free mitigation we recommend deploying alongside any AO used to surface specific factual claims.

\begin{figure}[htbp]
  \centering
  \includegraphics[width=0.6\linewidth]{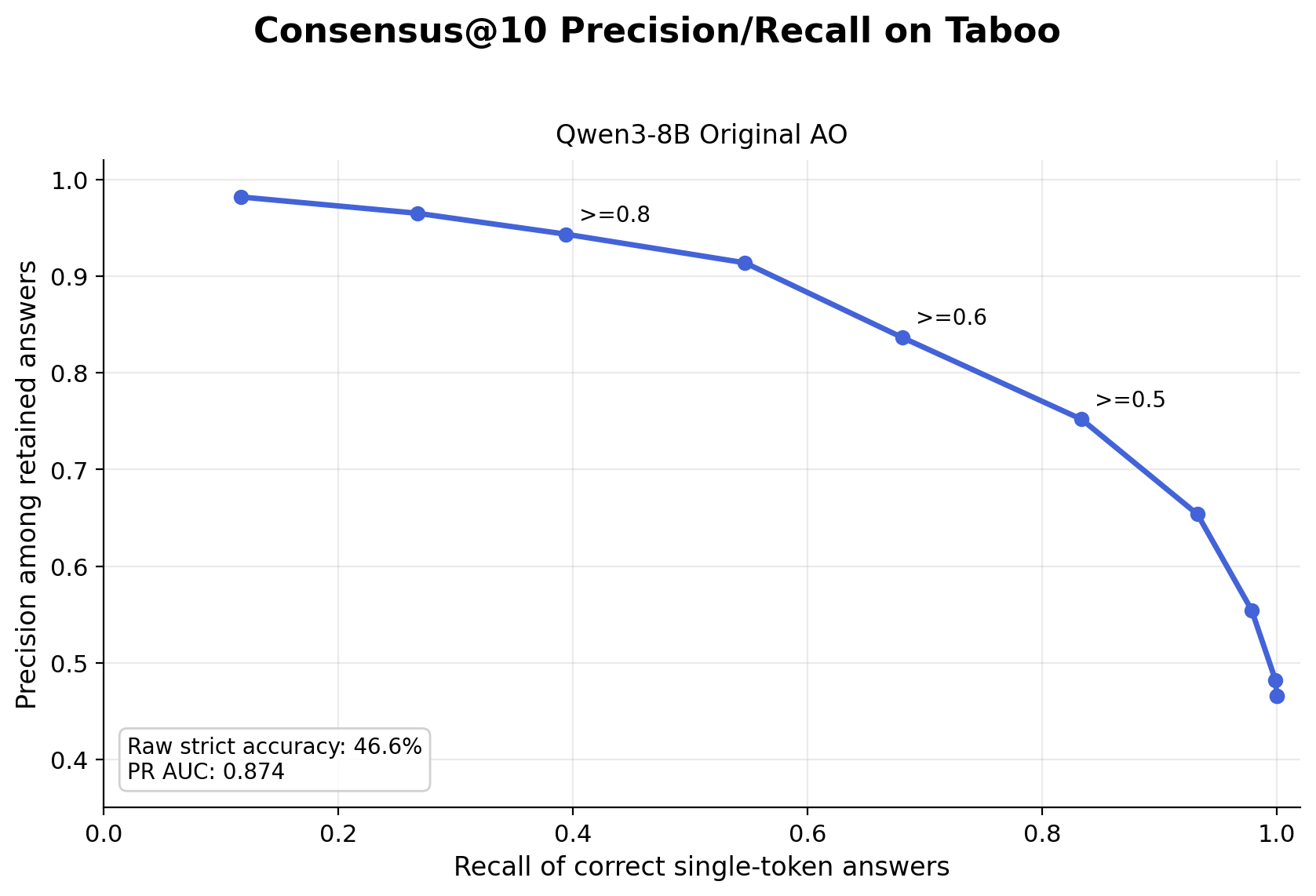}
  \caption{\textbf{Consensus@10 precision/recall on Taboo.} Requiring agreement among $k=10$ samples cleanly trades coverage for precision, mitigating hallucination on the secret-word extraction task.}
  \label{fig:consensus}
\end{figure}

\subsection{Our advice for training Activation Oracles}
\label{app:hillclimb}

Our initial impression was that we could improve AOs by training on narrow tasks. Specifically, we singled out the tasks from ``\href{https://www.lesswrong.com/posts/tDJWZLQNN7poqCwKa/test-your-best-methods-on-our-hard-cot-interp-tasks}{Test your best methods on our hard CoT interp tasks}'' (datasets can be found \href{https://huggingface.co/collections/mats-10-sprint-cs-jb/cleaned-datasets}{here}). We found that we could quite consistently match the performance of linear probes when narrowly training, but never significantly exceeded it.

\begin{itemize}
  \item Make a good eval, that you think a good Activation Oracle should be able to do (\textbf{solvability}) and is hard for a black box monitor (\textbf{text inversion}; you can explicitly check this)\footnote{A thing we tried to do during our sprint, was training a model that does not get activations, but everything in the context window up until that point (the same data the activations get to attend to), trained on the same objective. The reason this might work is that, because AOs are the same model, this model would actually be able to internally access these same activations. We did run this baseline at one point early in our sprint, and performance was similar, but we are sufficiently uncertain that we do not feel comfortable sharing this result as strong evidence. If you were able to demonstrate this matches AO performance on all tasks, AOs would still be useful as an interpretability tool (because the statement ``does this specific activation contain this information'' is still interesting, or for auditing different models), but this would mean they are not more useful as a monitorability tool.}. Then try to find training tasks that would make the oracle better at this.
  \item You should generally aim to match the performance of probes.\footnote{It was a very consistent observation that we were able to match linear probe performance, but never significantly exceed it.}
  \item A good training task causes broad uplift, and is scalable.
  \item Loss graphs going down does not always translate to capability: in particular, future/pastlens demonstrates a very strong scaling law, but there is a risk of just fitting surface statistics that will not translate to any meaningful uplift in evals.
  \item We observed the majority of uplift on the evals after 10\% of training (\textasciitilde{}200K tokens); training to convergence is generally not necessary to detect whether a task causes uplift.
  \item Be careful when changing learning rate, LoRA rank or LoRA alpha, as they can destabilize training.
  \item We experimented with scheduling training tasks one by one (unshuffled) to locate uplift, but encountered catastrophic forgetting on tasks not included in the group. Therefore, we recommend you have at least 10\% of data at every stage come from other tasks. An interesting way forward would be to have a broad ``pre-training stage'', say of verbatim and conversational data, and then a shorter ``post-train'' on specialized tasks.\footnote{A thing we experimented with was trying to do 2 epochs: 1 where the data is 90\% future/past-lens and 10\% conversational, to teach capability, and then a second epoch which was 90\% conversational, and 10\% future/past-lens, to hopefully make it answer more conversational, thus increasing specificity/reducing hallucinations. This did not lead to meaningful uplift.}
  \item Read your datasets, oracle outputs, and evaluation traces. Language models are not very good at generating or discriminating good AO questions/responses, so manual inspection helps verify the pipeline is behaving as intended.
\end{itemize}

\paragraph{Future directions we are excited about.}
\begin{itemize}
  \item Increasing corpus diversity on the unsupervised learning task.\footnote{Our chain-of-thought corpus consisted disproportionately of math, which is probably not optimal.}
  \item Feeding even more, or all, layers and positions.
  \item Training directly on activations from finetuned models, to optimise for model-diffing tasks.
\end{itemize}

\subsection{Feeding multiple positions}

We observed that hallucinations are reduced by feeding activations from more positions to the AO. The link was approximately linear: the more activations are fed, the fewer hallucinations. This makes sense, since the relevant information may be spread out across many activations and need not be concentrated in any single one.

\subsection{Experiments using post-training}

We experimented with DPO after final training to improve on instruction-following, hallucination rate, and vagueness. Results were hard to stabilize, and we frequently ran into mode-collapse. We found it hard to make an LLM judge correctly label ``good'' or ``bad'' Activation Oracle outputs, even with an explicit rubric. We also attempted GRPO-style RL with the following rubric:

\begin{itemize}
  \item Passing the ``swap test'' (does the AO's answer change when given activations from a meaningfully different context?)
  \item Is the answer specific and falsifiable?
  \item Does the response add any meaningful insight?
  \item Is it not clearly, obviously wrong?
  \item Is the oracle following instructions?
\end{itemize}

This increased performance on some evals but caused regressions on others, and we are not confident the resulting checkpoint is more useful to researchers in practice. We are uncertain whether our implementation choices were optimal given time constraints. We nonetheless believe RL remains a promising direction for aligning AO outputs with desirable properties; the central bottleneck is getting an LLM judge to reliably discriminate good from bad AO responses.

\subsection{\texorpdfstring{Other differences compared to \citet{karvonen2025activation}}{Other differences compared to Karvonen et al.}}

\begin{itemize}
  \item The number and identity of activation positions we feed is a random subset of the available positions (since our input is long). Adam's recipe uses ``sometimes contiguous $n$, sometimes a single activation''. We ablated this difference and did not find a significant change in eval performance.
\end{itemize}

\subsection{AObench details}
\label{app:aobench}

\begin{figure}[htbp]
  \centering
  \includegraphics[width=\linewidth]{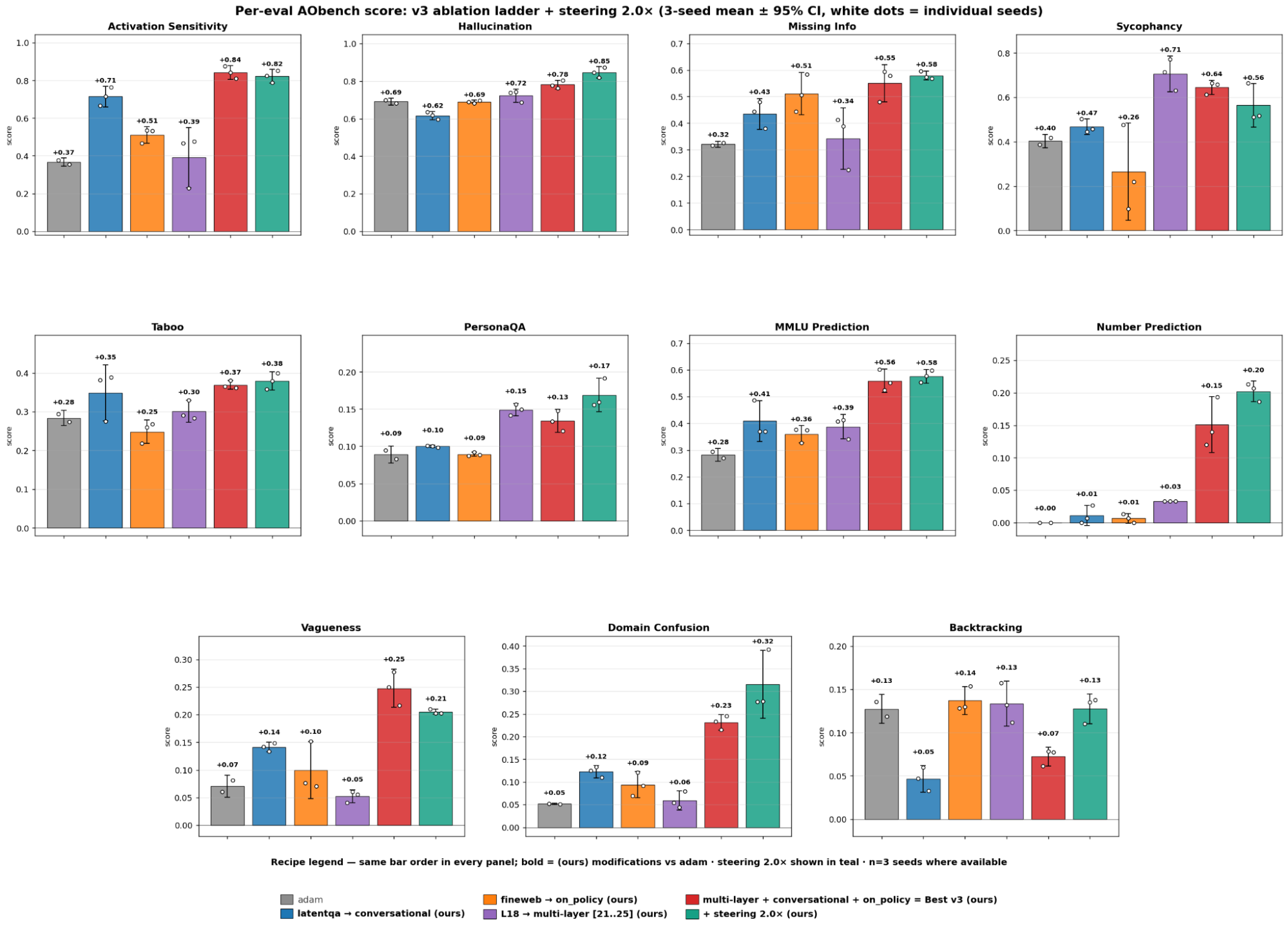}
  \caption{\textbf{Per-eval AObench scores across the ablation ladder.} Bars show chance-adjusted scores per task for each recipe in our ablation; white dots are individual seeds. Higher is better for every eval. The broad uplift visible in \cref{fig:ablation-waterfall} is reflected across the majority of individual tasks, with no single task driving the result.}
  \label{fig:aobench-results}
\end{figure}

AObench comprises the following individual evaluations, each chosen to target a capability an ideal AO should possess while attempting to remain robust to text-inversion confounds:

\begin{itemize}
  \item \textbf{Activation Sensitivity} (``Not Just Reading Tokens''). Feeds the exact same set of tokens but sourced from different upstream contexts, and checks whether the AO produces meaningfully different answers. This provides a relatively clean test of whether the AO relies on activation-specific information rather than surface token patterns.
  \item \textbf{Hallucination} (``Not Obviously Wrong''). Checks whether the AO is obviously confabulating given the context the activation comes from, i.e.\ whether it invents specific but unsupported details about the model's reasoning.
  \item \textbf{Missing Information} (``Detect Missing Info''). Adapted from \citet{jakkli2026current}, using SimpleBench-style questions where the model lacks a key piece of information but does not recognise this gap. Tests whether the oracle can surface knowledge that is not explicitly present in the reasoning.
  \item \textbf{Sycophancy}. Based on the Scruples dataset, this task assesses whether the oracle can detect sycophantic responses from the underlying activations produced during generation.
  \item \textbf{Taboo} (``Detect Taboo''). A model-organism task derived from \citet{cywinski2025elicit} and reused from \citet{karvonen2025activation}: the oracle must identify a secret word that the model deliberately avoids mentioning.
  \item \textbf{PersonaQA} (``Identify Persona''). Questions inquiring about the system prompt in use, approximating a steering vector; reused from \citet{karvonen2025activation}. Tests the oracle's knowledge of the instructions shaping the model's behaviour.
  \item \textbf{MMLU Prediction}. Requires predicting the answer the model would give to an MMLU question if it were interrupted mid-reasoning and forced to respond immediately.
  \item \textbf{Number Prediction} (``Predict Hidden Number''). Predicts the numbers involved in an arithmetic equation from the activations of that equation, testing numerical extraction.
  \item \textbf{Vagueness} (inverse of ``Response Specificity''). Measures whether the oracle's description of the model's reasoning is concrete, problem-specific and falsifiable, as opposed to a generic, unfalsifiable statement.
  \item \textbf{Domain Confusion} (``Identify Problem Domain''). Measures how often the model is completely wrong about the domain the activation is from (e.g.\ confusing ice cubes with hen houses), testing basic contextual awareness.
  \item \textbf{Backtracking}. Predicts whether the model is about to revise its current line of reasoning, which requires reading internal uncertainty signals.
\end{itemize}

The hallucination score reported in \cref{fig:hallucination-ladder} is a normalised average of: ``Not Obviously Wrong'', ``Identify Problem Domain'', ``Detect Missing Info'', and ``Predict Hidden Number''. Vagueness in \cref{fig:vagueness-ladder} is the inverse of ``Response Specificity''.

``Not Just Reading Tokens'' is a particularly promising eval, where we observe significant uplift: it feeds the exact same set of tokens but in different upstream contexts and checks whether the AO produces meaningfully different answers, providing a relatively clean test of whether the AO uses activation-specific information. ``Identify Persona'' and ``Detect Taboo'' are taken from \citet{karvonen2025activation}.

Our evaluation tasks are open-sourced at \href{https://github.com/ceselder/cot-oracle}{cot-oracle}. We reiterate that evaluating AOs is hard, chiefly due to the need to control for text inversion, and that judging vagueness reliably requires careful prompt design and manual spot-checking. We recommend qualitative analysis of every new AO checkpoint in addition to AObench.

\end{document}